\documentclass[11pt]{article}
\usepackage{coling2020}
\usepackage{times}
\usepackage{url}
\usepackage{latexsym}
\usepackage{booktabs}
\usepackage[ruled,vlined]{algorithm2e}
\usepackage{graphicx}
\graphicspath{{images/}}
\usepackage{caption}
\usepackage{subcaption}
\usepackage{amsmath}
\usepackage{amsfonts}
\usepackage{multirow}

\colingfinalcopy 

\title{Second-Order Unsupervised Neural Dependency Parsing}

\author{Songlin Yang\textsuperscript{1,2,3}, Yong Jiang\textsuperscript{4}, Wenjuan Han\textsuperscript{5}, Kewei Tu\textsuperscript{1}\thanks{\, 	 Corresponding Author}\\
  \textsuperscript{1}School of Information Science and Technology, ShanghaiTech University \\
  \textsuperscript{2}{Shanghai Institute of Microsystem and Information Technology, Chinese Academy of Sciences}\\
  \textsuperscript{3}{University of Chinese Academy of Sciences}\\
\textsuperscript{4}Alibaba DAMO Academy, Alibaba Group\\
  \textsuperscript{5}Department of Computer Science, National University of Singapore \\
    {\tt \{yangsl,tukw\}@shanghaitech.edu.cn}\\
    {\tt yongjiang.jy@alibaba-inc.com}\qquad
    {\tt dcshanw@nus.edu.sg}\\
 }

\date{}

\begin{document}
\maketitle
\begin{abstract}
Most of the unsupervised dependency parsers are based on first-order probabilistic generative models that only consider local parent-child information. Inspired by second-order supervised dependency parsing, we proposed a second-order extension of unsupervised neural dependency models  that incorporate grandparent-child or sibling information.  We also propose novel design of the neural parameterization and optimization methods of the dependency models. In second-order models, the number of grammar rules grows cubically with the increase of vocabulary size, making it difficult to train lexicalized models that may contain thousands of words. To circumvent this problem while still benefiting from both second-order parsing and lexicalization, we use the agreement-based learning framework to jointly train a second-order unlexicalized model and a first-order lexicalized model.   Experiments on multiple datasets show the effectiveness of our second-order models compared with recent state-of-the-art methods. Our joint model achieves a 10\% improvement over the previous state-of-the-art parser on the full WSJ test set.\footnote{Our source code is available at: https://github.com/sustcsonglin/second-order-neural-dmv}
\end{abstract}

\section{Introduction}
\label{intro}
%
%

Dependency parsing is a classical task in natural language processing. The head-dependent relations produced by dependency parsing can provide an approximation to the semantic relationship between words, which is useful in many downstream NLP tasks such as machine translation, information extraction and question answering.  Nowadays, supervised dependency parsers can reach a very high accuracy \cite{DBLP:conf/iclr/DozatM17,DBLP:conf/acl/ZhangLZ20}. Unfortunately, supervised parsing requires treebanks (annotated parse trees) for training, which are very expensive and time-consuming to build. On the other hand, unsupervised dependency parsing requires only unannotated corpora for training, though the accuracy of unsupervised parsing still lags far behind that of supervised parsing.  We focus on unsupervised dependency parsing in this paper. 

Most methods in the literature of unsupervised dependency parsing are based on the Dependency Model with Valence (DMV) \cite{Klein2004CorpusBasedIO}, which is a probabilistic generative model. A main disadvantage of DMV and many of its extensions is that they lack expressiveness. The generation of a dependent token is only conditioned on its parent, the relative direction of the token to its parent, and whether its parent has already generated any child in this direction, hence ignoring other contextual information. To improve model expressiveness, researchers often turn to discriminative methods, which can incorporate more contextual information into the scoring or prediction of dependency arcs. For example, \newcite{Grave2015ACA} uses the idea of disrciminative clustering,  \newcite{Cai2017CRFAF} uses a discriminative parser in the CRF-autoencoder framework, and \newcite{Li2018DependencyGI} uses an encoder-decoder framework that contains a discriminative transitioned-based parser. For DMV, \newcite{Han2019EnhancingUG} proposes the discriminative neural DMV which uses a global sentence embedding to introduce contextual information into the calculation of grammar rule probabilities. In the literature of supervised graph-based dependency parsing, however, there exists another technique for incorporating contextual information and increasing expressiveness, namely high-order parsing \cite{Koo2010EfficientTD,Ma2012FourthOrderDP}. A first-order parser, such as the DMV, only considers local parent-children information. In comparison, a high-order parser takes into account the interaction between multiple dependency arcs.

In this work, we propose the second-order neural DMV model, which incorporates second-order information (e.g., sibling or grandparent) into the original (neural) DMV model. To achieve better learning accuracy, we design a new neural architecture for rule probability computation and promote direct marginal likelihood optimization \cite{Salakhutdinov2003OptimizationWE,tran-etal-2016-unsupervised} over the widely used expectation-maximization algorithm for training. One particular challenge faced by second-order neural DMVs is that the number of grammar rules grows cubically to the vocabulary size, making it difficult to store and train a lexicalized model containing thousands of words. Therefore, instead of learning a second-order lexicalized model, we propose to jointly learn a second-order unlexicalized model (whose vocabulary consists of POS tags instead of words) and a first-order lexicalized model based on the agreement-based learning framework \cite{Liang2007AgreementBasedL}. The jointly learned models have a manageable number of grammar rules while still benefiting from both second-order parsing and lexicalization.

We conduct experiments on the Wall Street Journal (WSJ) dataset and seven languages on the Universal Dependencies (UD) dataset. The experimental results demonstrate that our models achieve state-of-the-art accuracies on unsupervised dependency parsing. 

    

\section{Background}

\subsection{Dependency Model With Valence}
The Dependency Model with Valence (DMV) \cite{Klein2004CorpusBasedIO} is a probabilistic generative model of a sentence and its parse tree.  It generates a dependency parse tree from the imaginary root node in a recursive top-down manner. There are three types of probabilistic grammar rules in a DMV, namely ROOT, CHILD and DECISION rules, each associated with a set of multinomial distributions $P_{\text{ROOT}}(c)$, $P_{\text{CHILD}}(c|p,dir,val)$ and $P_{\text{DECISION}}(dec|p,dir,val)$, where $p$ is the parent token, $c$ is the child token, $dec$ is the continue/stop decision, $dir$ indicates the direction of generation, and $val$ indicates whether parent $p$ has generated any child in direction $dir$. To generate a sequence of tokens along with its dependency parse tree, the DMV model generates a token $c$ from the ROOT distribution $
P_{\text{ROOT}}(c)$ firstly. Then for each token $p$ that has already been generated, it generates a decision from the DECISION distribution $P_{\text{DECISION}}(dec|p,dir,val)$ to determine whether to generate a new child in direction $dir$. If $dec$ is CONTINUE, then a new child $p$ is generated from the CHILD distribution $P_{\text{CHILD}}(c|p,dir,val)$. If $dec$ is STOP, then $p$ stops generating children in direction $dir$. The joint probability of the sequence and its corresponding dependency parse tree can be calculated by taking product of the probabilities of all the generation steps.

\subsection{Neuralized DMV Models}
\paragraph{Neural DMV} One limitation of the DMV model is that it does not consider the correlation between tokens.  \newcite{Jiang2016UnsupervisedND} proposed the Neural DMV (NDMV) model, which uses continuous POS embedding to represent discrete POS tags and calculate rule probabilities through neural networks based on the POS embedding. In this way, the model can learn the correlation between POS tags and smooth grammar rule probabilities accordingly.

\paragraph{Lexicalized NDMV}
Neural DMV is still an unlexicalized model which is based on POS tags and does not use word information. \newcite{Han2017DependencyGI} proposed the Lexicalized NDMV (L-NDMV) in which each token is a POS/word pair. The neural network that computes rule probabilities takes both the POS embedding and the word embedding as input. To reduce the vocabulary size, they replace low-frequency words with their POS tags. 

\section{Method}
\subsection{Second-Order Parsing}
In our proposed second-order NDMV, we calculate each rule probability based additionally on the information of the sibling or grandparent. We take sibling-NDMV for example to demonstrate the generative story.
\begin{itemize}
\setlength{\itemsep}{0pt}
\setlength{\parsep}{0pt}
\setlength{\parskip}{0pt}
    \item We start with the imaginary root token, generating its only child $c$ with probability  $P_{\text{ROOT}}(c)$
    \item For each token $p$, we decide whether to generate a new child or not with probability $P_{\text{DECISION}}(dec|p, s, dir, val)$, where $s$ is the previous child token generated by $p$ in direction $dir$.  If $p$ has not generated any child in direction $dir$ yet, we use a special symbol
   NULL to represent $s$. 
   \item If decision $dec$ is CONTINUE, $p$ generates a new child $c$ with probability $P_{\text{CHILD}}(c|p, s, dir,val)$. If decision $a$ is STOP, $p$ stops generating children in direction $dir$.
\end{itemize}

For parsing, we design dynamic programming algorithms adapted from \newcite{Koo2010EfficientTD}. Since the grandparent token is deterministic for each token, the parsing algorithm of our grand-NDMV model is similar to theirs. There are two options for determining the sibling token since the generation process of child tokens can be either from the inside out or from the outside in. \newcite{Koo2010EfficientTD} make the inside-out assumption, 
but in this paper, we make the outside-in assumption 
because it makes implementation easier and can achieve better performance empirically. We provide the pseudo code of the second-order inside algorithm and the second-order parsing algorithm in the appendix.

\subsection{Parameterization}
\label{section: parameterization}
\begin{figure}[ht]   
    \centering  
    
    \includegraphics[width=\textwidth ]{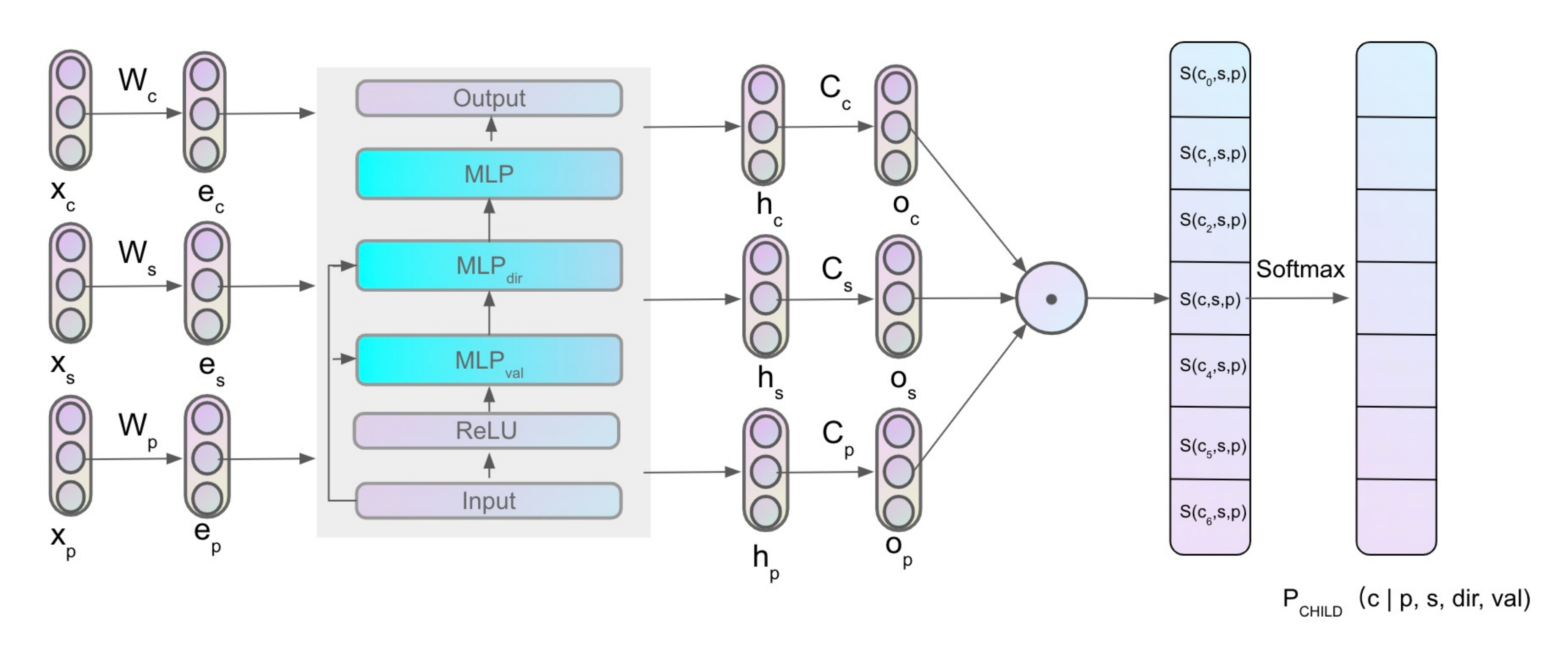}   
    \caption{Illustration of our neural architecture.}   
    \label{neural architecture}  
\end{figure}

In a neural DMV, we compute the probability of a grammar rule using a neural network. Below we formulate the computation of CHILD rule probabilities. The full architecture of the neural network is shown in Figure \ref{neural architecture}.
ROOT and DECISION rule probabilities are computed in a similar way.

In our second-order neural DMV, each CHILD rule $P_{\text{CHILD}}(c|p,s,dir,val)$ involves three tokens: parent $p$, child $c$, and sibling (or grandparent) $s$. 
Denote the embedding of the parent, child and sibling (or grandparent) by $x_{p}, x_{c}, x_{s} \in \mathcal{R}^{d}$, which are retrieved from a shared token embedding layer. We use three different linear transformations to produce the representations of a token as a parent, child, and sibling (or grandparent).
\[ e_{c} = W_{c} x_{c} \quad e_{p} = W_{p} x_{p}  \quad e_{s} = W_{s} x_{s} \]
We feed $e_{c}, e_{p}, e_{s}$ to the same neural network that consists of three consecutive MLPs. The first and second MLPs are used respectively to insert valence and direction information into the representations, and the last MLP is used to produce final hidden representations $h_{c}, h_{p}, h_{s}$ (see the appendix for the complete formulation). We use different parameters of the first and second MLPs for different values of valence $val$ and direction $dir$. We add skip-connections to the first and second MLPs because skip-connections have been found very useful in unsupervised neural parsing \cite{DBLP:conf/acl/KimDR19}. 

We then follow \newcite{Wang2019SecondOrderSD} and use a decomposed trilinear function to compute the unnormalized rule probability from the three vectors $h_{c}, h_{p}, h_{s}$. 
\[ S(p, s, c) = \sum\limits_{i=1}^{q} o_{p, i} \times o_{s, i} \times o_{c, i} \] 
\[ o_{p} = C_{p} h_{p} \quad o_{c} = C_{c} h_{c} \quad o_{s} = C_{s} h_{s} \]
where $C_{p},C_{c},C_{s} \in R^{q \times d}$ are the parameters of the decomposed trilinear function and $\times$ is scalar multiplication.
Then we apply a softmax function to produce the final rule probability. 
\[ P_{\text{CHILD}}(c | p, s, dir, val) = \frac{e^{S(p, s, c)}}{\sum\limits_{c^{\prime} \in \mathcal{C}} e^{S(p, s, c^{\prime})}} \]
where $\mathcal{C}$ is the vocabulary.

\subsection{Learning}
\label{dmo}
The learning objective function $L(\theta)$ is the log-likelihood of training sentences $ X = \{ x_{1},...,x_{n} \}$
\begin{equation}
L(\theta) = \sum\limits_{i=1}\limits^{n}\log p_{\theta}( x_{i})
\end{equation}   
where $\theta$ is the parameters of the neural networks. The probability of each sentence $x$ is defined as:
\begin{equation}
    p_{\theta}(x)=\sum_{z \in \mathcal{T}(x)} p_{\theta}(x, z)
\label{objective}
\end{equation}
where $\mathcal{T}(x)$ is the set of all possible dependency parse trees for sentence $x$.  We use $c(r,x,z)$ to represent the number of times rule $r$ is used in dependency parse tree $z$ of sentence $x$. Then we have 
\begin{equation}
p_{\theta}(x, z)=\prod_{r \in \mathcal{R}} p_{\theta}(r)^{c(r, x, z)}
\end{equation}
where $\mathcal{R}$ is the collection of all DECISION, CHILD and ROOT rules.

\paragraph{Learning via EM algorithm}
We can rewrite the log-likelihood of sentence $x$ as follows.
\begin{equation}
\log p_{\theta}(x) =   
E_{q(z)} 
\left[ \log p_{\theta}(x ,z) \right] + H\left[q(z)\right] + KL( q(z)  \| p_{\theta}(z |  x))
\end{equation}
where $q(z)$ is an arbitrary distribution and $H$ is the entropy function. In the E-step, we fix $\theta$ and set $q(z) = p_\theta(z|x)$. In the M-step, we fix $q(z)$ and update $\theta$ with the following objective:
\begin{equation}
Q(\theta) = E_{q(z)} \left[ \log p_{\theta}(x,z) \right] = E_{q(z)} \left[ \sum\limits_{r \in \mathcal{R}} c(r,x,z) \log p_{\theta}(r)  \right] = \sum\limits_{r \in \mathcal{R}} e(r,x) \log p_{\theta}(r)
\end{equation}
where $e(r,x)$ is the expected count of grammar rule $r$ in sentence $x$ based on $q(z)$, which can be obtained using the inside-outside algorithm. We can use gradient descent to update $\theta$. 
\begin{equation}
\setlength\abovedisplayskip{3pt}
\setlength\belowdisplayskip{3pt}
     \nabla_{\theta} Q(\theta) =  \sum_{r \in \mathcal{R}} e(r,x) \nabla_{\theta} \log p_{\theta}(r)
\label{M-step gradient}
\end{equation}

\paragraph{Learning via direct marginal likelihood optimization} 
We can also use gradient descent to maximize $\log p_{\theta}(x)$ directly. Based on the derivation of \newcite{Salakhutdinov2003OptimizationWE}\footnote{See Equation 8 in \newcite{Salakhutdinov2003OptimizationWE}.}, we have 
\begin{equation}
\setlength\abovedisplayskip{3pt}
\setlength\belowdisplayskip{3pt}
\begin{aligned}
\nabla_{\theta}(\log p_{\theta}(x))
&= \sum\limits_{z \in \mathcal{T}(x)} p_{\theta}(z |  x) \nabla_{\theta} \log p_{\theta}(x, z)  \\
&= \sum\limits_{z \in \mathcal{T}(x)} p_{\theta}(z |  x)
\sum\limits_{r \in \mathcal{R}} c(r, x, z) \nabla_{\theta} \log p_{\theta}(r) \\ 
&= \sum\limits_{r \in \mathcal{R}} e(r,x) \nabla_{\theta} \log p_{\theta}(r)
\end{aligned}
\label{direct marginal optimization}
\end{equation}
where $e(r,x)$ is the expected count of grammar rule $r$ in sentence $x$ based on $p_{\theta}(z|x)$. Traditionally, we use the inside-outside algorithm to obtain the expected count $e(r,x)$. \newcite{Eisner2016InsideOutsideAF} points out that we can use back-propagation to calculate the expected count $e(r,x)$.
\begin{equation}
 e(r,x) = \frac{\partial \log p_{\theta}(x)}{\partial \log p_{\theta}(r)} 
\end{equation}
So we only need to use the inside algorithm to calculate $\log p_{\theta}(x)$ and then use back-propagation to update the parameters directly, without the need for the outside algorithm. 

\paragraph{Mini-batch gradient descent as online EM}
In Equation \ref{direct marginal optimization}, we note that the gradient contains the term $e(r,x)$. If we use mini-batch gradient descent to optimize $\log p_{\theta}(x)$, it is analogous to the online-EM algorithm \cite{Liang2009OnlineEF}. To compute the gradient for each mini-batch, we first need to compute the expected counts from the training sentences in the mini-batch, which is exactly what the online E-step does; we then use the expected counts  to compute the gradient and update the model parameters, which is similar to the M-step, except that here we only perform one update step, while in the EM algorithm multiple update steps may be taken based on the same expected counts. According to \newcite{Liang2009OnlineEF}, online-EM has a faster convergence speed and can even find a better solution. Empirically, we do find that direct marginal likelihood optimization outperforms the EM algorithm.

\subsection{Agreement-Based Learning}
In our second-order DMV model, the number of grammar rules is $4\left|V\right|^3 + 4\left|V\right|^2 + \left|V\right|$, which is cubic in the vocabulary size $|V|$. When our model is lexicalized, the vocabulary may contain thousands of words or more, making the model size less manageable. Instead of learning a second-order lexicalized model, we propose to jointly learn a second-order unlexicalized model (whose vocabulary consists of POS tags instead of words) and a first-order lexicalized model based on the agreement-based learning framework \cite{Liang2007AgreementBasedL}. The jointly learned models have a manageable number of grammar rules while still benefiting from both second-order parsing and lexicalization. Empirically, we do find that the jointly trained models outperform lexicalized second-order models. 
 
Following \newcite{Liang2007AgreementBasedL}, we define the objective function for our jointly trained first-order L-NDMV and second-order NDMV as
\begin{equation}
    \mathcal{O}_{agree}(\theta) \stackrel{  def }{=} \log \sum_{z \in \mathcal{T}(x)} ( p_{\theta_{0}}(x, z) \cdot p_{\theta_{1}}(x, z) )
\end{equation}
where $\theta_{0}$ is parameters of L-NDMV and $\theta_{1}$ is parameters of second-order NDMV. Intuitively, the objective requires the two models to reach agreement on the probability distribution of dependency parse tree $z$. 
We use joint decoding (parsing) to predict dependency parse tree $z_{\text{predict}}$ for sentence $x$.
\begin{equation}
      z_{\text{predict}} = \operatorname{argmax}\limits_{z \in \mathcal{T}(x)} p_{\theta_{0}}(x,z) \cdot p_{\theta_{1}}(x,z) 
\end{equation} 
The inside and parsing algorithms for jointly trained models can be found in the appendix.

\paragraph{Learning via product EM algorithm} 
\newcite{Liang2007AgreementBasedL} propose to optimize the objective using the product EM algorithm based on the following lower bound of the objective.
\begin{equation}
\mathcal{O}_{agree}=\log \sum_{z} q(z) \frac{p_{\theta_{0}}\left( x, z \right) \cdot  p_{\theta_{1}}\left( x, z \right)}{q(z)} \ge E_{q(z)}(\log \frac{p_{\theta_{0}}\left( x, z \right) \cdot  p_{\theta_{1}}\left( x, z \right)}{q(z)})  \stackrel{  def }{=} L(\theta, q)
\end{equation}
The product EM algorithm performs coordinate-wise ascent on $L(\theta, q)$.   In the product E-step, we optimize $L(\theta, q)$ with respect to $q$.
 \begin{equation}
    L(\theta, q) = -KL(q(z) \| p_{\theta_{0}}\left( x, z \right) \cdot  p_{\theta_{1}}\left( x, z \right) ) + const 
 \end{equation} 
where $const$ does not depend on $\theta$ and $q$.  In the product E-step, the maximum can be obtained  by setting $q(z) \propto p_{\theta_{0}}\left( x, z \right) \cdot  p_{\theta_{1}}\left( x, z \right) $ to minimize the KL term.
In the M-step, we optimize $L(\theta,q)$ with respect to $\theta$.
\begin{equation}
L(\theta, q) = E_{q} \log p_{\theta_{0}}\left( x, z \right)  + E_{q} \log p_{\theta_{1}}\left( x, z \right) + const    
\end{equation}
where $const$ does not depend on $\theta$. It consists of one term for each model. We update the parameters of each model separately based on the expected counts obtained from the product E-step, which can be calculated through the inside-outside algorithm.

\paragraph{Learning via direct marginal likelihood optimization}  $\mathcal{O}_{agree}$ can be calculated through the inside algorithm. Similar to what we describe in Section \ref{dmo}, we can benefit from both agreement-based learning and the online-EM algorithm if we use gradient descent directly to optimize $\mathcal{O}_{agree}$ instead of using the product EM algorithm.

\section{Experiment}
\subsection{Datasets and Setting}
\paragraph{English Penn Treebank}  We conduct experiments on the Wall Street Journal (WSJ) corpus, with section 2-21 for training, section 22 for validation and section 23 for testing.  We use sentences of length $\le$ 10 in training and use sentences of length $\le$ 10 (WSJ10) and all sentences (WSJ) in testing.

\paragraph{Universal Dependency Treebank} Following the setting of \newcite{Li2018DependencyGI} and \newcite{Han2019EnhancingUG}, we conduct experiments on selected languages from the Universal Dependency Treebank
1.4 \cite{Nivre2016UniversalDV}. We use sentences of length $\le$ 15 in training and sentences of length $\le$ 15 and $\le$ 40 in testing. 

\paragraph{Setting} On the WSJ dataset, for fair comparison, we follow \newcite{Han2017DependencyGI} and \newcite{Han2019EnhancingUG} and use HDP-DEP \cite{Naseem2010UsingUL} to initialize our models. Specifically, we train the unsupervised HDP-DEP model on WSJ, use it to parse the training corpus, and then use the predicted parse trees to perform supervised learning of our model for several epochs. On the UD dataset, we use the K\&M initialization \cite{Klein2004CorpusBasedIO}.
We use direct marginal likelihood optimization (DMO) as the training method and use Adam \cite{Kingma2015AdamAM} as the optimizer with learning rate 0.001. The batch size is set to 64 for WSJ and 100 for UD.
The hyperparameters of the neural networks, the setting of L-NDMV and more details can be found in the appendix. We apply early stopping based on the log-likelihood of the development data and report the mean accuracy over 5 random restarts.

\subsection{Result}
\paragraph{Result on WSJ}In Table \ref{wsjresult}, we compare our methods with previous unsupervised dependency parsers. Our sibling-NDMV model can outperform the previous state-of-the-art parser by 1.9 points on WSJ10 and 3.1 points on WSJ in the unlexicalized setting.  Our lexicalized sibling-NDMV achieves further improvement over the unlexicalized sibling-NDMV. On the other hand, our grand-NDMV performs significantly worse than the sibling-NDMV and lexicalization hurts its performance. Why grandparent information is less useful than sibling information in unsupervised parsing is an intriguing question that we leave for feature research. 
Joint training with a first-order L-NDMV can increase the performance of unlexicalized sibling-NDMV from 77.5 to 79.9 and that of unlexicalized grand-NDMV from 71.4 to 76.0 on WSJ10. The jointly trained models also outperform the lexicalized second-order models.

\paragraph{Result on UD} In Table \ref{ud_result}, we first compare our models with models which do not use the universal linguistic prior (UP)\footnote{The universal linguistic prior is a set of syntactic dependencies that are common in many languages, proposed by \newcite{Naseem2010UsingUL}}. The variational variant of D-NDMV \cite{Han2019EnhancingUG}
is the recent state-of-the-art model without UP. Our method outperforms theirs on six of the eight languages and also on average. We then compare our second-order models with recent state-of-the-art discriminative models, which rely heavily on the universal linguistic prior to achieve good performance (for example, \newcite{Li2018DependencyGI} reported bad results if they do not use the universal linguistic prior). We find that sibling-NDMV can outperform these discriminative models while grand-NDMV can achieve comparable results, even though we do not utilize the universal linguistic prior.

\begin{table}[tbp]
\centering
\small
\begin{tabular}{cccc}
\hline
\multicolumn{2}{c|}{\bf\textsc{Methods}}              & \multicolumn{1}{c|}{\bf\textsc{WSJ10}} & \bf\textsc{WSJ}  \\ \hline
\multicolumn{2}{c|}{DMV\cite{Klein2004CorpusBasedIO}}                  & \multicolumn{1}{c|}{58.3}  & 39.4 \\
\multicolumn{2}{c|}{LN \cite{cohen2009logistic}}                   & \multicolumn{1}{c|}{59.4}  & 40.5 \\
\multicolumn{2}{c|}{Convex-MST \cite{Grave2015ACA}}           & \multicolumn{1}{c|}{60.8}  & 48.6 \\
\multicolumn{2}{c|}{Shared LN \cite{Cohen2009SharedLN}}            & \multicolumn{1}{c|}{61.3}  & 41.4 \\
\multicolumn{2}{c|}{Feature DMV  \cite{berg-kirkpatrick-etal-2010-painless}}          & \multicolumn{1}{c|}{63.0}  & -    \\
\multicolumn{2}{c|}{PR-S \cite{Gillenwater2011PosteriorSI}}                 & \multicolumn{1}{c|}{64.3}  & 53.3 \\
\multicolumn{2}{c|}{E-DMV \cite{Headden2009ImprovingUD}}                & \multicolumn{1}{c|}{65.0}  & -    \\
\multicolumn{2}{c|}{TSG-DMV \cite{Blunsom2010UnsupervisedIO}}              & \multicolumn{1}{c|}{65.9}  & 53.1 \\
\multicolumn{2}{c|}{UR-A E-DMV \cite{Tu2012UnambiguityRF}}           & \multicolumn{1}{c|}{71.4}  & 57.0 \\
\multicolumn{2}{c|}{CRFAE \cite{Cai2017CRFAF}}                & \multicolumn{1}{c|}{71.7}  & 55.7 \\
\multicolumn{2}{c|}{Neural DMV \cite{Jiang2016UnsupervisedND}}         & \multicolumn{1}{c|}{72.5}  & 57.6 \\
\multicolumn{2}{c|}{HDP-DEP \cite{Naseem2010UsingUL}}               & \multicolumn{1}{c|}{73.8}  & -    \\
\multicolumn{2}{c|}{NVTP \cite{Li2018DependencyGI}}                 & \multicolumn{1}{c|}{54.7}  & 37.8 \\
\multicolumn{2}{c|}{Variational variant D-NDMV * \cite{Han2019EnhancingUG}}                          & \multicolumn{1}{c|}{75.5} & 60.4 \\
\multicolumn{2}{c|}{Deterministic variant D-NDMV * \cite{Han2019EnhancingUG}}                        & \multicolumn{1}{c|}{75.6} & 61.4 \\
\multicolumn{2}{c|}{L-NDMV * \cite{Han2017DependencyGI}}         & \multicolumn{1}{c|}{75.1}  & 59.5 \\ \hline
\multicolumn{2}{c|}{grand-NDMV *}                              & \multicolumn{1}{c|}{71.4} & 57.3 \\
\multicolumn{2}{c|}{sibling-NDMV *}                              & \multicolumn{1}{c|}{77.5} & 64.5 \\
\multicolumn{2}{c|}{Lexicalized grand-NDMV *} & \multicolumn{1}{c|}{63.0}      &  52.6    \\ 
\multicolumn{2}{c|}{Lexicalized sibling-NDMV *} & \multicolumn{1}{c|}{78.3}      & 66.4     \\ \hline
\multicolumn{2}{c|}{Joint training: grand-NDMV + L-NDMV *}   &
\multicolumn{1}{c|}{76.0}     &   64.3   \\
\multicolumn{2}{c|}{Joint training: sibling-NDMV + L-NDMV *}  &
\multicolumn{1}{c|}{ \textbf{79.9}}     &  \textbf{67.5}    \\ \hline
\multicolumn{4}{c}{Systems with Additional Training Data (for reference)}                                   \\ \hline
\multicolumn{2}{c|}{CS \cite{Spitkovsky2013BreakingOO}}                   & \multicolumn{1}{c|}{72.0}  & 64.4 \\
\multicolumn{2}{c|}{MaxEnc \cite{le-zuidema-2015-unsupervised}}               & \multicolumn{1}{c|}{73.2}  & 65.8 \\
\multicolumn{2}{c|}{L-NDMV * \cite{Han2017DependencyGI}}               & \multicolumn{1}{c|}{77.2}  & 63.2 \\ \hline
\end{tabular}
\caption{ Comparison on WSJ. *: Approaches which use  \newcite{Naseem2010UsingUL} for initialization.}
\label{wsjresult}
\end{table}

\begin{table}[tbp]
\centering
\small
\begin{tabular}{ccccccccc}
\hline
\multicolumn{1}{c|}{} &
  \multicolumn{6}{c|}{\bf\textsc{NO UP}} &
  \multicolumn{2}{c}{+\bf\textsc{UP}} \\ \cline{2-9} 
\multicolumn{1}{c|}{} &
  NDMV &
  LD &
  DV &
  \multicolumn{1}{c|}{VV} &
  +sibling &
  \multicolumn{1}{c|}{+grand} &
  NVTP &
  CM \\ \hline
\multicolumn{9}{c}{Length $\le$ 15} \\ \hline
\multicolumn{1}{c|}{Basque} &
  48.3 &
  47.9 &
  40.6 &
  \multicolumn{1}{c|}{42.7} &
  30.8 &
  \multicolumn{1}{c|}{34.1} &
  \textbf{52.9} &
  52.5 \\
\multicolumn{1}{c|}{Dutch} &
  44.1 &
  35.5 &
  42.1 &
  \multicolumn{1}{c|}{43.0} &
  \textbf{46.5} &
  \multicolumn{1}{c|}{42.4} &
  39.6 &
  43.4 \\
\multicolumn{1}{c|}{French} &
  59.5 &
  52.1 &
  59.0 &
  \multicolumn{1}{c|}{61.7} &
  \textbf{65.7} &
  \multicolumn{1}{c|}{62.9} &
  59.9 &
  61.6 \\
\multicolumn{1}{c|}{German} &
  56.2 &
  51.9 &
  56.4 &
  \multicolumn{1}{c|}{58.5} &
  \textbf{63.6} &
  \multicolumn{1}{c|}{48.9} &
  57.5 &
  66.7 \\
\multicolumn{1}{c|}{Italian} &
  72.7 &
  73.1 &
  59.6 &
  \multicolumn{1}{c|}{63.5} &
  \textbf{76.3} &
  \multicolumn{1}{c|}{76.3} &
  59.7 &
  73.2 \\
\multicolumn{1}{c|}{Polish} &
  72.7 &
  66.2 &
  70.5 &
  \multicolumn{1}{c|}{75.8} &
  \textbf{77.1} &
  \multicolumn{1}{c|}{63.7} &
  57.1 &
  66.7 \\
\multicolumn{1}{c|}{Portuguese} &
  45.5 &
  \textbf{70.5} &
  68.8 &
  \multicolumn{1}{c|}{69.1} &
  67.8 &
  \multicolumn{1}{c|}{62.5} &
  52.7 &
  60.7 \\
\multicolumn{1}{c|}{Spanish} &
  38.1 &
  65.5 &
  63.8 &
  \multicolumn{1}{c|}{66.1} &
  67.2 &
  \multicolumn{1}{c|}{\textbf{67.5}} &
  55.6 &
  61.6 \\ \hline
\multicolumn{1}{c|}{\textbf{Average}} &
  53.3 &
  57.8 &
  57.6 &
  \multicolumn{1}{c|}{60.1} &
  \textbf{61.9} &
  \multicolumn{1}{c|}{57.2} &
  54.4 &
  59.3 \\ \hline
\multicolumn{9}{c}{Length $\le$ 40} \\ \hline
\multicolumn{1}{c|}{Basque} &
  47.8 &
  45.4 &
  39.9 &
  \multicolumn{1}{c|}{42.4} &
  30.5 &
  \multicolumn{1}{c|}{33.0} &
  48.9 &
  \textbf{50.0} \\
\multicolumn{1}{c|}{Dutch} &
  35.6 &
  34.1 &
  42.4 &
  \multicolumn{1}{c|}{43.7} &
  45.0 &
  \multicolumn{1}{c|}{43.7} &
  42.5 &
  \textbf{45.3} \\
\multicolumn{1}{c|}{French} &
  38.1 &
  48.6 &
  57.2 &
  \multicolumn{1}{c|}{58.5} &
  \textbf{64.9} &
  \multicolumn{1}{c|}{61.4} &
  55.4 &
  62.0 \\
\multicolumn{1}{c|}{German} &
  50.4 &
  50.5 &
  54.5 &
  \multicolumn{1}{c|}{52.9} &
  \textbf{61.5} &
  \multicolumn{1}{c|}{46.7} &
  54.2 &
  51.4 \\
\multicolumn{1}{c|}{Italian} &
  63.6 &
  71.1 &
  60.2 &
  \multicolumn{1}{c|}{61.3} &
  \textbf{71.8} &
  \multicolumn{1}{c|}{69.7} &
  55.7 &
  69.1 \\
\multicolumn{1}{c|}{Polish} &
  62.8 &
  63.7 &
  66.7 &
  \multicolumn{1}{c|}{73.0} &
  \textbf{75.0} &
  \multicolumn{1}{c|}{62.4} &
  51.7 &
  63.4 \\
\multicolumn{1}{c|}{Portuguese} &
  49.0 &
  \textbf{67.2} &
  64.7 &
  \multicolumn{1}{c|}{65.7} &
  63.7 &
  \multicolumn{1}{c|}{60.3} &
  45.3 &
  57.1 \\
\multicolumn{1}{c|}{Spanish} &
  58.0 &
  61.9 &
  64.3 &
  \multicolumn{1}{c|}{64.4} &
  \textbf{66.8} &
  \multicolumn{1}{c|}{65.0} &
  52.5 &
  61.9 \\ \hline
\multicolumn{1}{c|}{\textbf{Average}} &
  50.7 &
  55.3 &
  56.2 &
  \multicolumn{1}{c|}{57.7} &
  \textbf{59.9} &
  \multicolumn{1}{c|}{55.2} &
  50.8 &
  57.5 \\ \hline
\end{tabular}
\caption{Comparison on Universal Dependency Treebanks. No UP: System using the universal linguistic prior. +UP : Systems using the universal linguistic prior.  LD: LC-DMV \cite{Noji2015LeftcornerPF}. DV,VV: The deterministic and variational variants of D-NDMV \cite{Han2019EnhancingUG}. +sibling: Our second-order sibling-NDMV. +grand: Our second-order grand-NDMV.  NVTP: Neural variational transition-based
parser \cite{Li2018DependencyGI}. CM: Convex-MST \cite{Grave2015ACA}.}
\label{ud_result}
\end{table}

\section{Analysis}
\subsection{Effect of Skip-Connections}
From Table \ref{UD: SC and EM} and \ref{WSJ: SC and EM}, we find that using skip-connections can achieve higher log-likelihood and better parsing accuracy in most cases. On UD, the performance is much better when using skip-connections except on Basque. 

\begin{table}[t]
\centering
\small
\begin{tabular}{c|cc|c|cc|c}
\hline
\bf\textsc{Language}  & \multicolumn{3}{c|}{\bf\textsc{Log-likelihood}}          & \multicolumn{3}{c}{\bf\textsc{UAS ( $\le 15$ / $\le 40$)}}    \\ \cline{2-7} 
          & \multicolumn{2}{c|}{DMO}             & EM    & \multicolumn{2}{c|}{DMO}               & EM        \\ \cline{2-7} 
          & \multicolumn{1}{c|}{w. SC} & w.o. SC & w. SC & \multicolumn{1}{c|}{w. SC} & w.o. SC   & w. SC     \\ \hline
Basque    & -19.3                      & -19.7   & -20.0 & 30.8/30.5                  & 44.7/44.3 & 28.8/29.1 \\
Dutch     & -19.1                      & -19.4   & -19.3 & 46.5/45.0                  & 34.9/34.8 & 43.6/41.8 \\
French    & -19.8                      & -20.9   & -19.4 & 65.7/64.9                  & 47.2/50.1 & 55.0/54.5 \\
German    & -20.9                      & -22.4   & -21.2 & 63.6/61.5                  & 36.4/37.8 & 57.9/54.5 \\
Italian   & -16.7                      & -16.9   & -16.7 & 76.3/71.8                  & 65.6/58.1 & 71.0/65.2 \\
Polish    & -16.2                      & -16.9   & -17.0 & 77.1/75.0                  & 66.2/63.7 & 61.7/60.4 \\
Portugese & -17.9                      & -18.4   & -17.8 & 67.8/63.7                  & 60.0/59.1 & 57.7/51.9 \\
Spanish   & -20.8                      & -21.7   & -20.9 & 67.2/66.8                  & 49.6/48.8 & 61.4/58.1 \\ \hline
\end{tabular}
\caption{Effect of skip-connections and training methods on UD.}
\label{UD: SC and EM}
\end{table}

\begin{table}[t]
\centering
\small
\begin{tabular}{c|cc|c|cc|c}
\hline
  \bf\textsc{Models}                               & \multicolumn{3}{c|}{\bf\textsc{Log-likelihood}}          & \multicolumn{3}{c}{\bf\textsc{UAS ( WSJ10 / WSJ)}}           \\ \cline{2-7} 
                                      & \multicolumn{2}{c|}{DMO}             & EM    & \multicolumn{2}{c|}{DMO}               & EM        \\ \cline{2-7} 
                                      & \multicolumn{1}{c|}{w. SC} & w.o. SC & w. SC & \multicolumn{1}{c|}{w. SC} & w.o. SC   & w. SC     \\ \hline
sibling-NDMV                          & -17.3                      & -17.7   & -17.3 & 77.5/64.5                  & 75.4/63.1 & 77.7/64.5 \\
grand-NDMV                            & -16.9                      & -17.4   & -17.0 & 71.4/57.3                  & 68.4/55.1 & 72.7/61.8 \\
sibling-NDMV + L-NDMV & -53.2                      & -55.3   & -53.5 & 79.9/67.5                  & 78.2/64.7 & 79.1/66.5 \\
grand-NDMV + L-NDMV  & -53.8                      & -57.5   & -54.2 & 76.0/64.3                  & 73.7/60.7 & 75.2/65.3 \\ \hline
\end{tabular}
\caption{Effect of skip-connections and training methods on WSJ.}
\label{WSJ: SC and EM}
\end{table}

\subsection{Comparison of Training Methods}
In Table \ref{UD: SC and EM}, we find that the EM algorithm significantly underperforms DMO. On the other hand, Table \ref{WSJ: SC and EM} shows that the EM algorithm performs comparably to DMO on WSJ.
 
\begin{figure}[tbp] 
\centering
  \begin{minipage}[b]{0.45\linewidth}
    \centering
        \includegraphics[width=7cm]{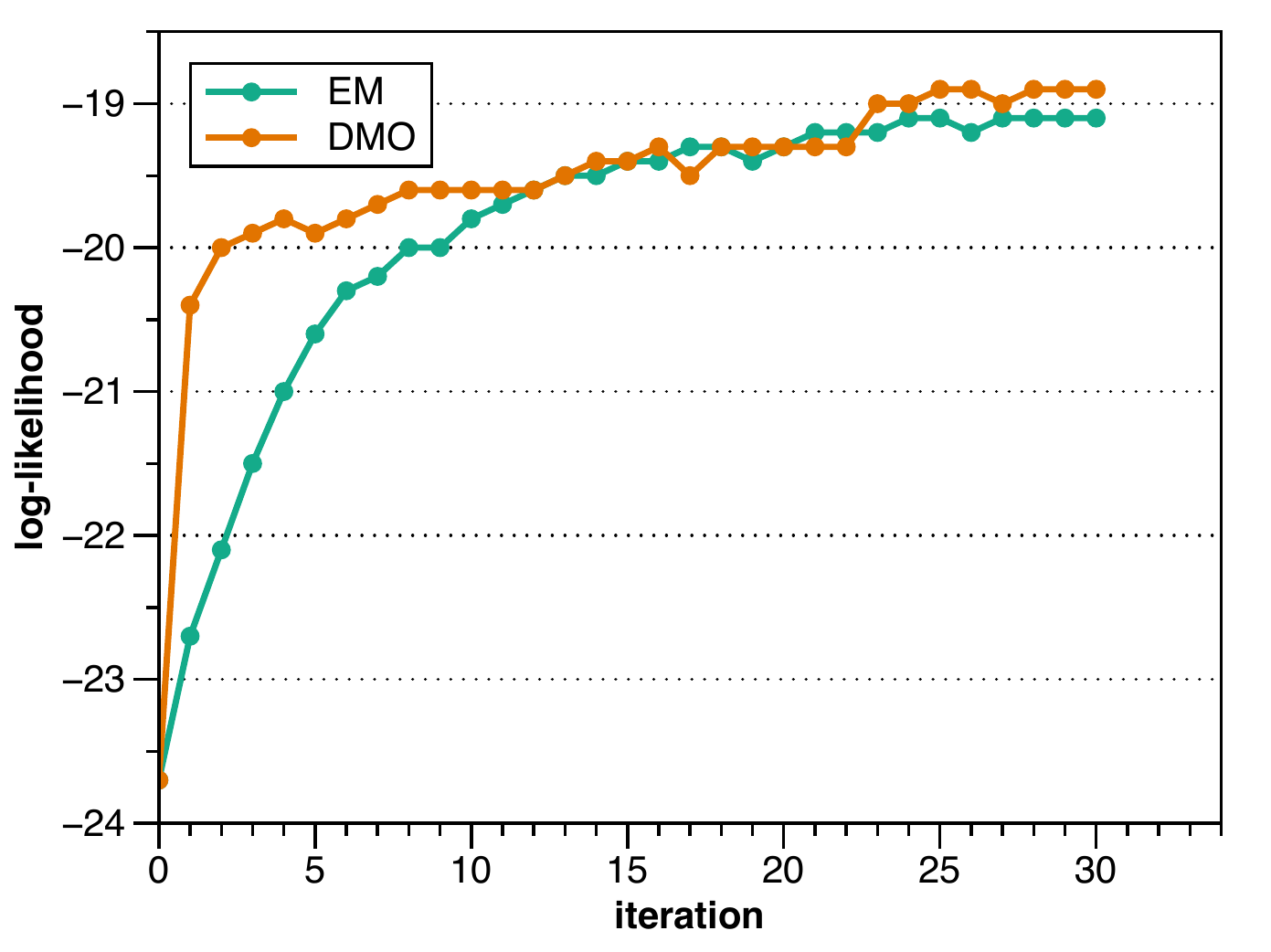} 
    \caption{Comparison of training methods on log-likelihood for WSJ.} 
          \label{wsj_likelihood} 
    \vspace{0ex}
  \end{minipage}
  \hspace*{2em}
  \begin{minipage}[b]{0.45\linewidth}
    \centering
    \includegraphics[width=7cm]{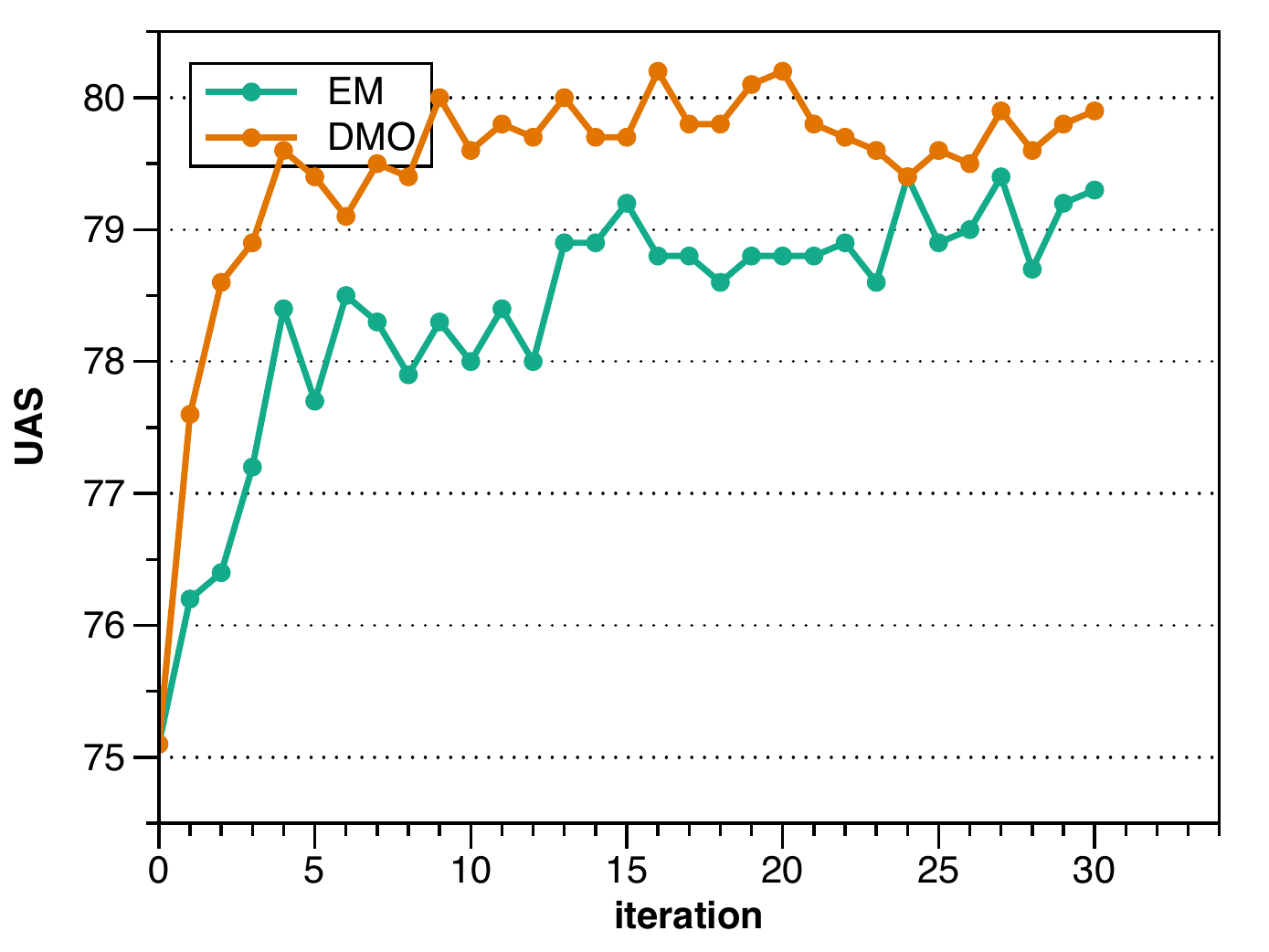} 
    \caption{Comparison of training methods on UAS for WSJ.} 
      \label{wsj_uas} 
    \vspace{0ex}
  \end{minipage} 
\end{figure}

\begin{figure}[tbp] 
\centering
  \begin{minipage}[b]{0.45\linewidth}
    \centering
    \includegraphics[width=7cm]{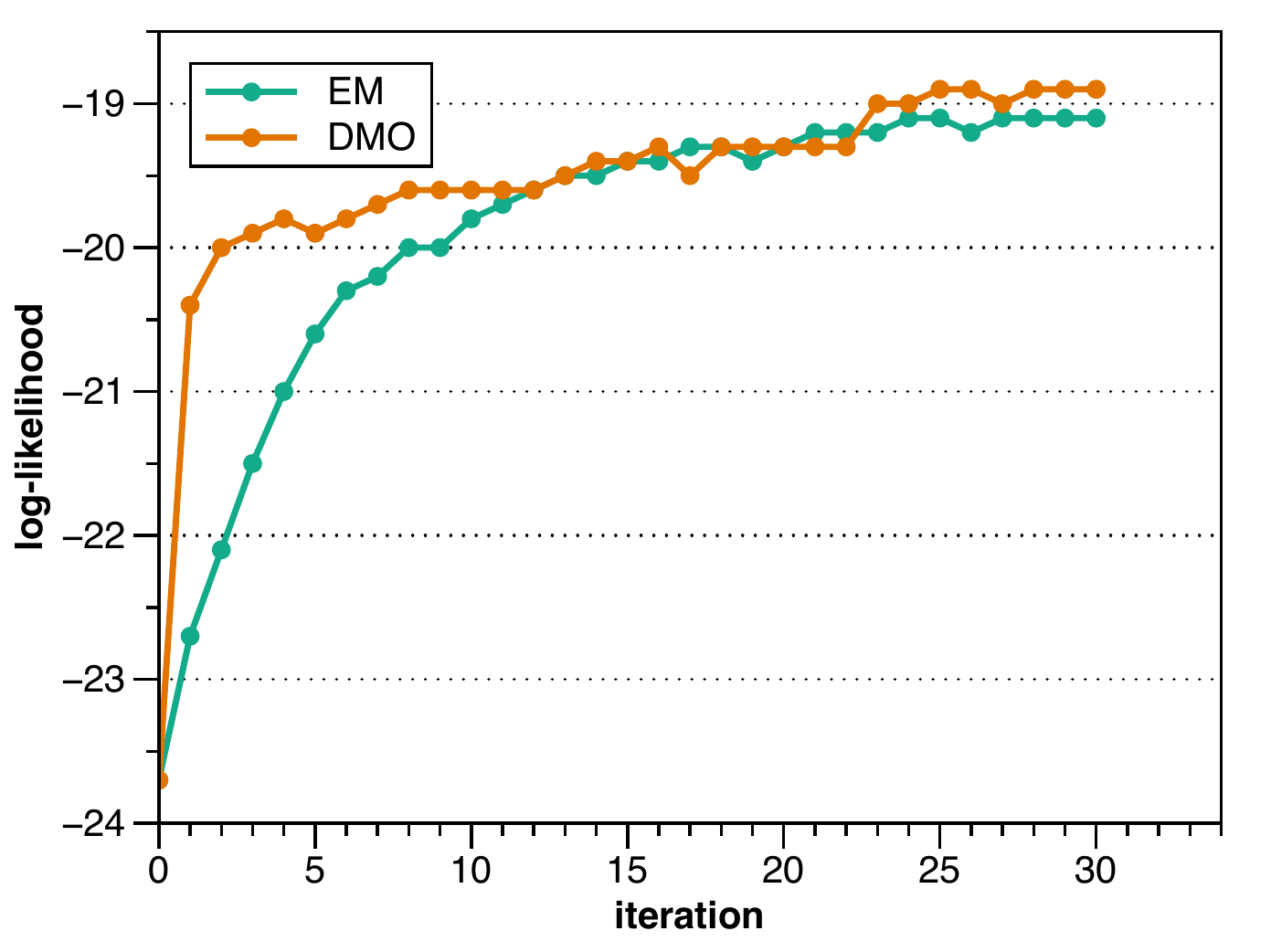}
    \caption{Comparison of training methods on log-likelihood for UD (French).}
    \label{ud-likelihood}
  \end{minipage}
  \hspace*{2em}
  \begin{minipage}[b]{0.45\linewidth}
    \centering
\includegraphics[width=7cm]{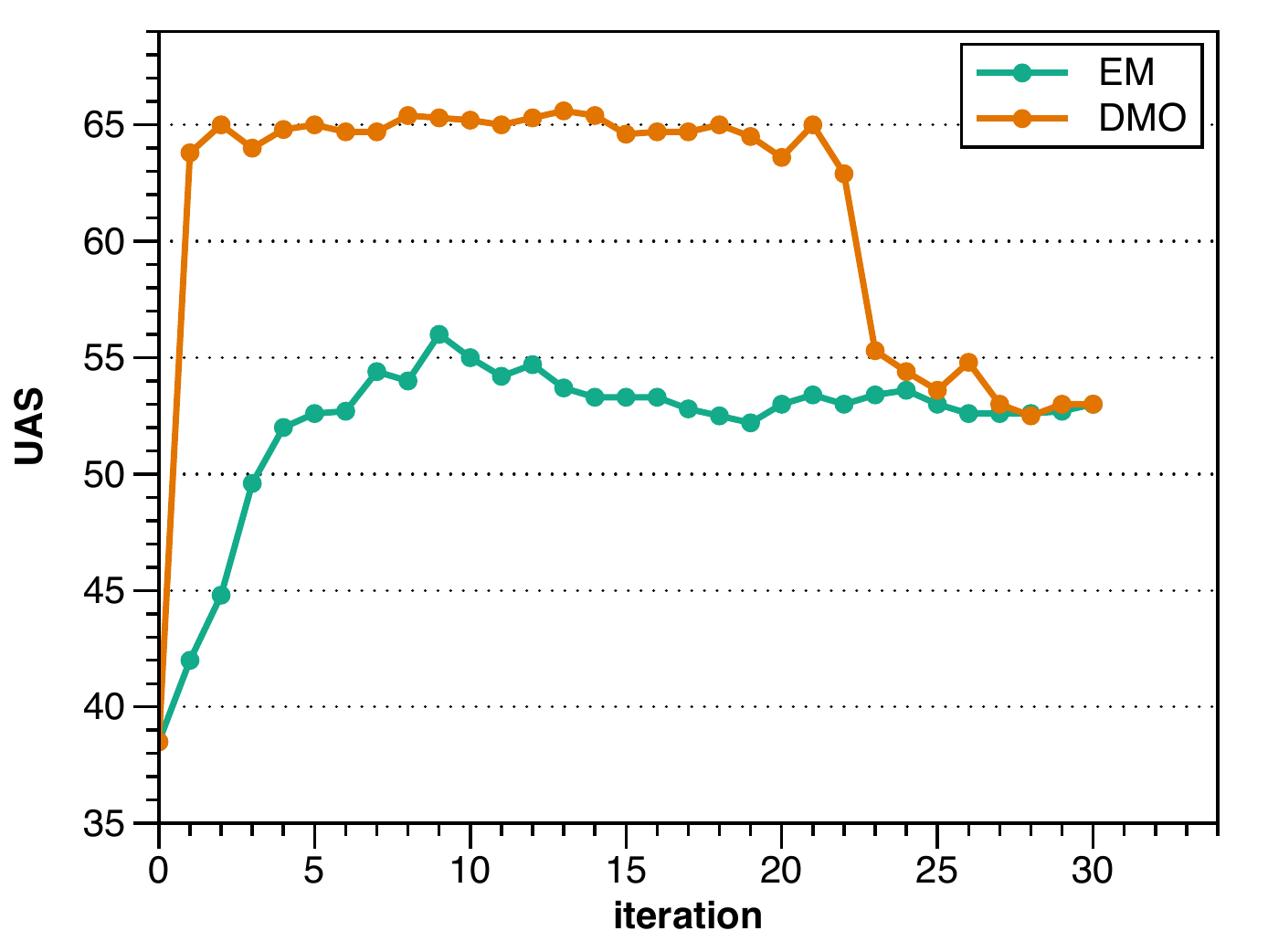} 
    \caption{Comparison of training methods on UAS for UD (French).} 
     \label{ud_uas} 
  \end{minipage} 
\end{figure}

We also compare the learning curves of these two methods. For fair comparison, we use the same batch-size for both methods. First we conduct an experiment using the joint L-NDMV and sibling-NDMV model on WSJ. In Figure \ref{wsj_likelihood}, we find that DMO converges to a higher log-likelihood compared with EM and the convergence speed is roughly the same.  In Figure \ref{wsj_uas}, we find DMO can find a slightly better model compared with EM. Second, we conduct an experiment using sibling-NDMV model on the UD French dataset. In Figure \ref{ud-likelihood}, we find DMO converges faster than EM and converges to a higher log-likelihood. In Figure \ref{ud-likelihood}, we find that the model accuracy of DMO is much higher than that of EM at the beginning, but it drops significantly after epoch 23, suggesting that early-stop is necessary. We also find similar phenomena for other languages on UD. 
 
It should be noted that we use HDP-DEP \cite{Naseem2010UsingUL} for initialization on WSJ and use K\&M initialization \cite{Klein2004CorpusBasedIO} on UD. We see that HDP-DEP initialization leads to a very high initial UAS of 75\% (Figure \ref{wsj_uas}),
while K\&M initialization leads to a low initial UAS of 38.5\% (Figure \ref{ud_uas}). It can be seen that EM is more sensitive to the initialization while DMO can achieve good results even if the initialization is bad. 

\begin{table}[tb]
\centering
\small
\begin{tabular}{l|l|l|l}
\hline
Training method   & \multicolumn{3}{l|}{UAS  (WSJ10 / WSJ)} \\ \hline
                  & L-NDMV         & sibling-NDMV   & joint parsing   \\ \hline
separate training & 76.6 / 62.7    & 77.5 / 64.8    & 78.4 / 65.8      \\ \hline
joint training    & 79.2 / 65.4    & 78.7 / 65.6    & 79.9 / 67.5      \\ \hline
\end{tabular}
\caption{The effect of joint training and joint parsing}
\label{joint-decoding}
\end{table}

\subsection{Effect of Joint Training and Parsing}
In Table \ref{joint-decoding}, we compare the performance with different training and parsing settings. We find that joint parsing is better than separate parsing in both training settings. With joint training, each individual model can achieve better performance compared with separate training, which shows the effectiveness of agreement-based joint learning.

\subsection{Limitations}
Our second-order NDMV model is more sensitive to the initialization compared with the first-order NDMV model. We fail to produce a good result under the K\&M initialization on WSJ: only 58.5\% UAS for sibling-NDMV on WSJ10, while the first-order NDMV model can achieve 69.7\% UAS. We rely on the parsing result of HDP-DEP to initialize our model in order to reach the state-of-the-art result on WSJ. This is similar to the case of L-NDMV, which performs badly when using the K\&M initialization according to \newcite{Han2017DependencyGI}. Because of the bad performance of L-NDMV with the K\&M initialization as well as the time constraint that prevents us from running HDP-DEP on UD, 
we did not conduct experiments of agreement-based learning with L-NDMV on the UD datasets. We leave this for future work. 
 
Our second-order model is also quite sensitive to the design of the neural architecture, which is similar to case of unsupervised constituency parsing reported by \newcite{DBLP:conf/acl/KimDR19}. We also try the third-order NDMV model (grand-sibling or tri-sibling) but are not able to get better results compared with sibling-NDMV. 

Our second-order parsing algorithm has a theoretical time complexity of $O(n^{4})$, which is higher than the time complexity of $O(n)$ of transition-based unsupervised parsers \cite{Li2018DependencyGI} and the time complexity of $O(n^{3})$ of first-order NDMV models, where $n$ is the sentence length. However, transition-based parsers are hard to batchify, while our model can be parallelized efficiently following the methods introduced by Torch-Struct \cite{rush-2020-torch}. In practice, our second-order parser runs very fast on GPU, requiring only several minutes to train.

\section{Conclusion}
We propose second-order NDMV models, which incorporate sibling or grandparent information. We find that sibling information is very useful in unsupervised dependency parsing. We use agreement-based learning to combine the benefits of second-order parsing and lexicalization, achieving state-of-the-art results on the WSJ dataset. We also show the effectiveness of our neural parameterization architecture with skip-connections and the direct marginal likelihood optimization method.

\section*{Acknowledgement}
This work was supported by the National Natural Science Foundation of China (61976139).

\bibliographystyle{coling}
\bibliography{coling2020}

\begin{thebibliography}{}

\bibitem[\protect\citename{Berg-Kirkpatrick \bgroup et al.\egroup
  }2010]{berg-kirkpatrick-etal-2010-painless}
Taylor Berg-Kirkpatrick, Alexandre Bouchard-C{\^o}t{\'e}, John DeNero, and Dan
  Klein.
\newblock 2010.
\newblock Painless unsupervised learning with features.
\newblock In {\em Human Language Technologies: The 2010 Annual Conference of
  the North {A}merican Chapter of the Association for Computational
  Linguistics}, pages 582--590, Los Angeles, California, June. Association for
  Computational Linguistics.

\bibitem[\protect\citename{Blunsom and Cohn}2010]{Blunsom2010UnsupervisedIO}
Phil Blunsom and Trevor Cohn.
\newblock 2010.
\newblock Unsupervised induction of tree substitution grammars for dependency
  parsing.
\newblock In {\em EMNLP}.

\bibitem[\protect\citename{Cai \bgroup et al.\egroup }2017]{Cai2017CRFAF}
Jiong Cai, Yong Jiang, and Kewei Tu.
\newblock 2017.
\newblock Crf autoencoder for unsupervised dependency parsing.
\newblock In {\em EMNLP}.

\bibitem[\protect\citename{Cohen and Smith}2009]{Cohen2009SharedLN}
Shay~B. Cohen and Noah~A. Smith.
\newblock 2009.
\newblock Shared logistic normal distributions for soft parameter tying in
  unsupervised grammar induction.
\newblock In {\em HLT-NAACL}.

\bibitem[\protect\citename{Cohen \bgroup et al.\egroup
  }2009]{cohen2009logistic}
Shay~B Cohen, Kevin Gimpel, and Noah~A Smith.
\newblock 2009.
\newblock Logistic normal priors for unsupervised probabilistic grammar
  induction.
\newblock In {\em Advances in Neural Information Processing Systems}, pages
  321--328.

\bibitem[\protect\citename{Dozat and Manning}2017]{DBLP:conf/iclr/DozatM17}
Timothy Dozat and Christopher~D. Manning.
\newblock 2017.
\newblock Deep biaffine attention for neural dependency parsing.
\newblock In {\em 5th International Conference on Learning Representations,
  {ICLR} 2017, Toulon, France, April 24-26, 2017, Conference Track
  Proceedings}. OpenReview.net.

\bibitem[\protect\citename{Eisner}2016]{Eisner2016InsideOutsideAF}
Jason Eisner.
\newblock 2016.
\newblock Inside-outside and forward-backward algorithms are just backprop
  (tutorial paper).
\newblock In {\em SPNLP@EMNLP}.

\bibitem[\protect\citename{Gillenwater \bgroup et al.\egroup
  }2011]{Gillenwater2011PosteriorSI}
Jennifer Gillenwater, Kuzman Ganchev, Jo{\~a}o Graça, Fernando~C Pereira, and
  Ben Taskar.
\newblock 2011.
\newblock Posterior sparsity in unsupervised dependency parsing.
\newblock {\em J. Mach. Learn. Res.}, 12:455--490.

\bibitem[\protect\citename{Grave and Elhadad}2015]{Grave2015ACA}
Edouard Grave and No{\'e}mie Elhadad.
\newblock 2015.
\newblock A convex and feature-rich discriminative approach to dependency
  grammar induction.
\newblock In {\em ACL}.

\bibitem[\protect\citename{Han \bgroup et al.\egroup
  }2017]{Han2017DependencyGI}
Wenjuan Han, Yong Jiang, and Kewei Tu.
\newblock 2017.
\newblock Dependency grammar induction with neural lexicalization and big
  training data.
\newblock In {\em EMNLP}.

\bibitem[\protect\citename{Han \bgroup et al.\egroup }2019]{Han2019EnhancingUG}
Wenjuan Han, Yong Jiang, and Kewei Tu.
\newblock 2019.
\newblock Enhancing unsupervised generative dependency parser with contextual
  information.
\newblock In {\em ACL}.

\bibitem[\protect\citename{Headden \bgroup et al.\egroup
  }2009]{Headden2009ImprovingUD}
William~P. Headden, Mark Johnson, and David McClosky.
\newblock 2009.
\newblock Improving unsupervised dependency parsing with richer contexts and
  smoothing.
\newblock In {\em HLT-NAACL}.

\bibitem[\protect\citename{Jiang \bgroup et al.\egroup
  }2016]{Jiang2016UnsupervisedND}
Yong Jiang, Wenjuan Han, and Kewei Tu.
\newblock 2016.
\newblock Unsupervised neural dependency parsing.
\newblock In {\em EMNLP}.

\bibitem[\protect\citename{Kim \bgroup et al.\egroup
  }2019]{DBLP:conf/acl/KimDR19}
Yoon Kim, Chris Dyer, and Alexander~M. Rush.
\newblock 2019.
\newblock Compound probabilistic context-free grammars for grammar induction.
\newblock In Anna Korhonen, David~R. Traum, and Llu{\'{\i}}s M{\`{a}}rquez,
  editors, {\em Proceedings of the 57th Conference of the Association for
  Computational Linguistics, {ACL} 2019, Florence, Italy, July 28- August 2,
  2019, Volume 1: Long Papers}, pages 2369--2385. Association for Computational
  Linguistics.

\bibitem[\protect\citename{Kingma and Ba}2015]{Kingma2015AdamAM}
Diederik~P. Kingma and Jimmy Ba.
\newblock 2015.
\newblock Adam: A method for stochastic optimization.
\newblock {\em CoRR}, abs/1412.6980.

\bibitem[\protect\citename{Klein and Manning}2004]{Klein2004CorpusBasedIO}
Dan Klein and Christopher~D. Manning.
\newblock 2004.
\newblock Corpus-based induction of syntactic structure: Models of dependency
  and constituency.
\newblock In {\em ACL}.

\bibitem[\protect\citename{Koo and Collins}2010]{Koo2010EfficientTD}
Terry~K Koo and Michael Collins.
\newblock 2010.
\newblock Efficient third-order dependency parsers.
\newblock In {\em ACL}.

\bibitem[\protect\citename{Le and Zuidema}2015]{le-zuidema-2015-unsupervised}
Phong Le and Willem Zuidema.
\newblock 2015.
\newblock Unsupervised dependency parsing: Let{'}s use supervised parsers.
\newblock In {\em Proceedings of the 2015 Conference of the North {A}merican
  Chapter of the Association for Computational Linguistics: Human Language
  Technologies}, pages 651--661, Denver, Colorado, May{--}June. Association for
  Computational Linguistics.

\bibitem[\protect\citename{Li \bgroup et al.\egroup }2018]{Li2018DependencyGI}
Bowen Li, Jianpeng Cheng, Yang Liu, and Frank Keller.
\newblock 2018.
\newblock Dependency grammar induction with a neural variational
  transition-based parser.
\newblock In {\em AAAI}.

\bibitem[\protect\citename{Liang and Klein}2009]{Liang2009OnlineEF}
Percy Liang and Dan Klein.
\newblock 2009.
\newblock Online em for unsupervised models.
\newblock In {\em HLT-NAACL}.

\bibitem[\protect\citename{Liang \bgroup et al.\egroup
  }2007]{Liang2007AgreementBasedL}
Percy Liang, Dan Klein, and Michael~I. Jordan.
\newblock 2007.
\newblock Agreement-based learning.
\newblock In {\em NIPS}.

\bibitem[\protect\citename{Ma and Hai}2012]{Ma2012FourthOrderDP}
Xuezhe Ma and Zhao Hai.
\newblock 2012.
\newblock Fourth-order dependency parsing.
\newblock In {\em COLING}.

\bibitem[\protect\citename{Naseem \bgroup et al.\egroup
  }2010]{Naseem2010UsingUL}
Tahira Naseem, Harr Chen, Regina Barzilay, and Mark Johnson.
\newblock 2010.
\newblock Using universal linguistic knowledge to guide grammar induction.
\newblock In {\em EMNLP}.

\bibitem[\protect\citename{Nivre \bgroup et al.\egroup
  }2016]{Nivre2016UniversalDV}
Joakim Nivre, Marie-Catherine de~Marneffe, Filip Ginter, Yoav Goldberg, Jan
  Hajic, Christopher~D. Manning, Ryan~T. McDonald, Slav Petrov, Sampo Pyysalo,
  Natalia Silveira, Reut Tsarfaty, and Daniel Zeman.
\newblock 2016.
\newblock Universal dependencies v1: A multilingual treebank collection.
\newblock In {\em LREC}.

\bibitem[\protect\citename{Noji and Miyao}2015]{Noji2015LeftcornerPF}
Hiroshi Noji and Yusuke Miyao.
\newblock 2015.
\newblock Left-corner parsing for dependency grammar.
\newblock {\em Journal of Information Processing}, 22:251--288.

\bibitem[\protect\citename{Rush}2020]{rush-2020-torch}
Alexander Rush.
\newblock 2020.
\newblock Torch-struct: Deep structured prediction library.
\newblock In {\em Proceedings of the 58th Annual Meeting of the Association for
  Computational Linguistics: System Demonstrations}, pages 335--342, Online,
  July. Association for Computational Linguistics.

\bibitem[\protect\citename{Salakhutdinov \bgroup et al.\egroup
  }2003]{Salakhutdinov2003OptimizationWE}
Ruslan Salakhutdinov, Sam~T. Roweis, and Zoubin Ghahramani.
\newblock 2003.
\newblock Optimization with em and expectation-conjugate-gradient.
\newblock In {\em ICML}.

\bibitem[\protect\citename{Spitkovsky \bgroup et al.\egroup
  }2013]{Spitkovsky2013BreakingOO}
Valentin~I. Spitkovsky, Hiyan Alshawi, and Dan Jurafsky.
\newblock 2013.
\newblock Breaking out of local optima with count transforms and model
  recombination: A study in grammar induction.
\newblock In {\em EMNLP}.

\bibitem[\protect\citename{Tran \bgroup et al.\egroup
  }2016]{tran-etal-2016-unsupervised}
Ke~M. Tran, Yonatan Bisk, Ashish Vaswani, Daniel Marcu, and Kevin Knight.
\newblock 2016.
\newblock Unsupervised neural hidden {M}arkov models.
\newblock In {\em Proceedings of the Workshop on Structured Prediction for
  {NLP}}, pages 63--71, Austin, TX, November. Association for Computational
  Linguistics.

\bibitem[\protect\citename{Tu and Honavar}2012]{Tu2012UnambiguityRF}
Kewei Tu and Vasant~G Honavar.
\newblock 2012.
\newblock Unambiguity regularization for unsupervised learning of probabilistic
  grammars.
\newblock In {\em EMNLP-CoNLL}.

\bibitem[\protect\citename{Wang \bgroup et al.\egroup
  }2019]{Wang2019SecondOrderSD}
Xinyu Wang, Jingxian Huang, and Kewei Tu.
\newblock 2019.
\newblock Second-order semantic dependency parsing with end-to-end neural
  networks.
\newblock In {\em ACL}.

\bibitem[\protect\citename{Zhang \bgroup et al.\egroup
  }2020]{DBLP:conf/acl/ZhangLZ20}
Yu~Zhang, Zhenghua Li, and Min Zhang.
\newblock 2020.
\newblock Efficient second-order treecrf for neural dependency parsing.
\newblock In Dan Jurafsky, Joyce Chai, Natalie Schluter, and Joel~R. Tetreault,
  editors, {\em Proceedings of the 58th Annual Meeting of the Association for
  Computational Linguistics, {ACL} 2020, Online, July 5-10, 2020}, pages
  3295--3305. Association for Computational Linguistics.

\end{thebibliography}




\appendix
\section{Appendix}
\subsection{Inside Algorithm and Parsing Algorithm}
\label{intro2}

We use the dynamic programming substructure proposed for second-order supervised dependency parsing.  For grandparent-child model, \newcite{Koo2010EfficientTD} augment both complete and incomplete spans with grandparent indices. They called the augmented span g-spans. Formally, they denote a complete g-span as $C_{h,e}^{g}$, where $C_{h,e}$ is a normal complete span in the Eisner algorithm, g is the grandparent's index, with the implication that $(g,h)$ is a dependency. Incomplete g-span is defined similarly. 

For second-order NDMV, we further augment incomplete and complete g-spans with valence information. We distinguish the direction of span explicitly, denoting our augmented complete v-span as $C_{h,e,d}^{g,v}$, where $d$ is the direction, $v$ is the valence, $h$ is the start index and $e$ is the end index of span compared with g-span. Incomplete v-span is defined similarly. 

For grand-NDMV, given sentence $x$, we suppose that $x_{0}$ is the imaginary root token and $x_{1}, .. x_{n}$ are tokens.
We denote $D[i, g, d, v, a] = \log(\rm p_{DECISION}(decision=a|parent=x_{i},grand=x_{g},direction=d, valence=v)) $, $S[i, c, g, d, v] = \log(\rm p_{CHILD}(child=x_{c}|parent=x_{i},grand=x_{g},direction=d,valence=v))$, and $R[i] = \log(\rm p_{ROOT}(child=x_{i}))$. Given these definitions, the inside algorithm of grand-NDMV is shown in Algorithm \ref{eisner grand}. 

For sibling-NDMV, $g$ in $C_{h,e}^{g}$ stands for the index of sibling instead of the index of grandparent. Given sentence $x$, we suppose that $x_{0}$ is a special NULL token which stands for no sibling and $x_{1}, .. x_{n}$ are tokens. We denote $D[i, g, d, v, a] = \log(\rm p_{DECISION}(decision=a|parent=x_{i},sibling=x_{g},direction=d, valence=v)) $, $S[i, c, g, d, v] = \log(\rm p_{CHILD}(child=x_{c}|parent=x_{i},sibling=x_{g},direction=d,valence=v))$, and $R[i] = \log(\rm p_{ROOT}(child=x_{i}))$. Given these definitions, the inside algorithm of sibling-NDMV is shown in Algorithm \ref{eisner sibling}.

For jointly trained L-NDMV and second-order NDMV model, we take jointly trained L-NDMV sibing-NDMV for example. We denote $D^{\prime}[i, d, v, a] = \log(\rm p_{DECISION}(decision=a|parent=x_{i}^{\prime},direction=d, valence=v)), S^{\prime}[i, c, d, v] = \log(\rm p_{CHILD}(child=x_{c}^{\prime}|parent=x_{i}^{\prime},direction=d,valence=v)), R^{\prime}[i] = \log(\rm p_{ROOT}(child=x_{i}^{\prime}))$ for L-NDMV where $x^{\prime}$ is the sequence of word/POS pairs which starts indexing at 1.
The inside algorithm of jointly trained L-NDMV and sibling-NDMV model is shown in Algorithm \ref{eisner joint}.
 
Following \newcite{Eisner2016InsideOutsideAF}, we use back-propagation to obtain expected counts of grammar rules.  For the parsing algorithm, we can replace $\text{logsumexp}$ with $\text{max}$ in Algorithm \ref{eisner grand}, \ref{eisner sibling} and \ref{eisner joint} to get the Viterbi log-likelihood of the sentence, then use back-propagation to get grammar rules which are used in the Viterbi parse tree, and finally reconstruct the parse tree based on these rules.

\subsection{Full Parameterization}

Denote the embedding of the parent, child and sibling (or grandparent) by $x_{p}, x_{c}, x_{s} \in \mathcal{R}^{d}$.  We use three different linear transformations to produce the representations of each token as a parent, child, and sibling (or grandparent).
\[ e_{c} = W_{c} x_{c} \quad e_{p} = W_{p} x_{p}  \quad e_{s} = W_{s} x_{s} \]

We feed $e_{s}, e_{c}, e_{p}$ to the same neural network which consists of three MLP with skip-connection layer. The first MLP aims at encoding valence information:
\[ v =  ReLU (W_{1}(ReLU(W_{val} e + e))) \] where $val \in  \left[ \rm HASCHILD, NOCHILD \right]$.

The second MLP aims at encoding direction information:
\[d =  W_{3}(ReLU(W_{2}(W_{dir} v + e)) \] where  $ dir \in \left[ \rm LEFT, RIGHT \right]$ 

We use the final MLP to get final hidden representation $h$:
\[h = ReLU(W_{4}d)\]

For the UD dataset, we use a more expressive MLP to get the final hidden representation $h$ since we find that the UD dataset is more difficult to train. 
\[h = ReLU(W_{6} (ReLU(W_{5}(ReLU(W_{4} d + e)))))\]

For decision rules, we introduce two embedding $x_{\text{stop}}$ and $x_{\text{continue}}$.  We feed $x_{\text{stop}}$ and $x_{\text{continue}}$ to a fully connected layer $W_{\text{dec}}$ to get $e_{\text{stop}}$ and $e_{\text{continue}}$. We use the same neural network to get hidden representation $h_{\text{stop}}$ and $h_{\text{continue}}$.  For root rules, we introduce $x_{\text{root}}$. We feed $x_{\text{root}}$ to a fully connected layer $W_{\text{root}}$ to get $e_{\text{root}}$. Also, we use the same neural network to get hidden representation $h_{\text{root}}$.  We use different decomposed trilinear function parameters for different types of rules. The calculation of the decision rule probability and root rule probability is similar to that of the child rule probability.

\subsection{Hyperparameters}
We set the dimension of POS embedding to 100. The dimension of all linear layers to calculate hidden representation is set to 100. We set the size of decomposed trilinear function parameters to 30 for child and root rules and 10 for decision rules in the unlexicalized setting. 

For the lexicalized model, we set the dimension of word embedding to 100. We concatenate the POS embedding and word embedding as input. The dimension of all linear layers to calculate hidden representation is set to 200.  We set the size of decomposed trilinear function parameters to 150 for child and root rules and 50 for decision rules. We use an additional dropout layer after the embedding layer to avoid over-fitting since the vocabulary size of the lexicalized model is much larger compared to the unlexicalized model. The dropout rate is set to 0.5.

\subsection{Setting of L-NDMV}
The vocabulary consists of word/POS pairs that appear for at least two times in the WSJ10 dataset. We use random embedding to initialize the POS embedding and FastText embedding to initialize the word embedding, which is different from the setting in the original paper \cite{Han2017DependencyGI}. We train FastText on the whole WSJ dataset for 100 epochs with window size 3 and embedding dimension 100. 
\begin{algorithm}[t]
\footnotesize
\SetAlgoLined
\textbf{notation} LEFT=0, RIGHT=0, HASCHILD=0, NOCHILD=1 \newline
\textbf{initialization}
 $ {\forall i,g}\quad$ $C_{i,i,0}^{g,0} = D[i,g,0,0,0]$
  $C_{i,i,0}^{g,1} = D[i,g,0,1,0]$
 $C_{i,i,1}^{g,0} = D[i,g,1,0,0]$ 
  $C_{i,i,1}^{g,1} = D[i,g,1,1,0]$ \newline
\bfseries{for} $w=1...(n-1)$ \newline
 \hspace*{1em} \bfseries{for} $i = 1 ... (n-w)$\newline
 \hspace*{2em} $j = i + w$ \newline
 \hspace*{2em} \bfseries{for} $g < i$ \bfseries{or}  $g > j$ \newline
 \hspace*{3em} $I_{i,j,0}^{g,0} = \boldsymbol{\rm logsumexp}_{i \le r <j} \left\{ C_{i,r,1}^{j,1} + C_{r+1,j,0}^{g,0} + D[j,g,0,0,1] + S[j,i,g,0,0] \right\} $ \newline
  \hspace*{3em} $I_{i,j,0}^{g,1} = \boldsymbol{\rm logsumexp}_{i \le r <j} \left\{ C_{i,r,1}^{j,1} + C_{r+1,j,0}^{g,0} + D[j,g,0,1,1] + S[j,i,g,0,1] \right\} $ \newline
   \hspace*{3em} $I_{i,j,1}^{g,0} = \boldsymbol{\rm logsumexp}_{i \le r <j} \left\{ C_{i,r,1}^{g,0} + C_{r+1,j,0}^{i,1} + D[i,g,1,0,1] + S[i,j,g,1,0] \right\} $ \newline
  \hspace*{3em} $I_{i,j,1}^{g,1} = \boldsymbol{\rm logsumexp}_{i \le r <j} \left\{ C_{i,r,1}^{g,0} + C_{r+1,j,0}^{i,1} + D[i,g,1,1,1] + S[i,j,g,1,1] \right\} $ \newline
     \hspace*{3em} $C_{i,j,0}^{g,0} = \boldsymbol{\rm logsumexp}_{i \le r \le j} \left\{ C_{i,r,0}^{j,1} + I_{r,j,0}^{g,0} \right\} $ \newline
     \hspace*{3em} $C_{i,j,0}^{g,1} = \boldsymbol{\rm logsumexp}_{i \le r \le j} \left\{ C_{i,r,0}^{j,1} + I_{r,j,0}^{g,1}  \right\} $ \newline
     \hspace*{3em} $C_{i,j,1}^{g,0} = \boldsymbol{\rm logsumexp}_{i \le r \le j} \left\{ I_{i,r,1}^{g,0} + C_{r,j,1}^{i,1} \right\} $ \newline
  \hspace*{3em} $C_{i,j,1}^{g,1} = \boldsymbol{\rm logsumexp}_{i \le r \le j} \left\{ I_{i,r,1}^{g,1} + C_{r,j,1}^{i,1}  \right\} $ \newline
\bfseries{for} $i = 1 .. (n)$\newline
\hspace*{1em} $P[i] = R[i] + C_{1,i,0}^{0,1}$ 

$P = \boldsymbol{\rm logsumexp}_{1\le i \le n} P[i] + C_{i,n,1}^{0,1}$ \newline
\Return P

 \caption{Inside algorithm for grand-NDMV}
 \label{eisner grand}
\end{algorithm}
\begin{algorithm}[t]
 \footnotesize
\SetAlgoLined
\textbf{notation} LEFT=0, RIGHT=0, HASCHILD=0, NOCHILD=1 \newline
 \textbf{initialization}
 $ {\forall i,g}\quad$ $C_{i,i,0}^{g,0} = D[i,g,0,0,0]$
  $C_{i,i,0}^{g,1} = D[i,g,0,1,0]$
 $C_{i,i,1}^{g,0} = D[i,g,1,0,0]$ 
  $C_{i,i,1}^{g,1} = D[i,g,1,1,0]$ \newline
\bfseries{for} $w=1...(n-1)$ \newline
 \hspace*{1em} \bfseries{for} $i = 1 ... (n-w)$\newline
 \hspace*{2em} $j = i + w$ \newline
 \hspace*{2em} \bfseries{for} $g < i$ \bfseries{or}  $g > j$ \newline
 \hspace*{3em} $I_{i,j,0}^{g,0} = \boldsymbol{\rm logsumexp}_{i \le r <j} \left\{ C_{i,r,1}^{0,1} + C_{r+1,j,0}^{i,0} + D[j,g,0,0,1] + S[j,i,g,0,0] \right\} $ \newline
  \hspace*{3em} $I_{i,j,0}^{g,1} = \boldsymbol{\rm logsumexp}_{i \le r <j} \left\{ C_{i,r,1}^{0,1} + C_{r+1,j,0}^{i,0} + D[j,g,0,1,1] + S[j,i,g,0,1] \right\} $ \newline
   \hspace*{3em} $I_{i,j,1}^{g,0} = \boldsymbol{\rm logsumexp}_{i \le r <j} \left\{ C_{i,r,1}^{j,0} + C_{r+1,j,0}^{0, 1} + D[i,g,1,0,1] + S[i,j,g,1,0] \right\} $ \newline
  \hspace*{3em} $I_{i,j,1}^{g,1} = \boldsymbol{\rm logsumexp}_{i \le r <j} \left\{ C_{i,r,1}^{j,0} + C_{r+1,j,0}^{0,1} + D[i,g,1,1,1] + S[i,j,g,1,1] \right\} $ \newline
     \hspace*{3em} $C_{i,j,0}^{g,0} = \boldsymbol{\rm logsumexp}_{i \le r \le j} \left\{ C_{i,r,0}^{0,1} + I_{r,j,0}^{i,0} \right\} $ \newline
     \hspace*{3em} $C_{i,j,0}^{g,1} = \boldsymbol{\rm logsumexp}_{i \le r \le j} \left\{ C_{i,r,0}^{0,1} + I_{r,j,0}^{i,1}  \right\} $ \newline
     \hspace*{3em} $C_{i,j,1}^{g,0} = \boldsymbol{\rm logsumexp}_{i \le r \le j} \left\{ I_{i,r,1}^{j,0} + C_{r,j,1}^{0,1} \right\} $ \newline
  \hspace*{3em} $C_{i,j,1}^{g,1} = \boldsymbol{\rm logsumexp}_{i \le r \le j} \left\{ I_{i,r,1}^{j,1} + C_{r,j,1}^{0,1}  \right\} $ \newline
\bfseries{for} $i = 1 .. (n)$ \newline
\hspace*{1em} $P[i] = R[i] + C_{1,i,0}^{0,1}$ \newline
$P = \boldsymbol{\rm logsumexp}_{1\le i \le n} P[i] + C_{i,n,1}^{0,1}$ \newline
\Return P
 \caption{Inside algorithm for sibling-NDMV} 
 \label{eisner sibling}
\end{algorithm}

\begin{algorithm}[t]
\footnotesize
\SetAlgoLined
\textbf{notation} LEFT=0, RIGHT=0, HASCHILD=0, NOCHILD=1 \newline
 \textbf{initialization}
 $ {\forall i,g}\quad$ $C_{i,i,0}^{g,0} = D[i,g,0,0,0] + D^{\prime}[i,0,0,0]$
  $C_{i,i,0}^{g,1} = D[i,g,0,1,0] + D^{\prime}[i,0,1,0]$
 $C_{i,i,1}^{g,0} = D[i,g,1,0,0] + D^{\prime}[i,1,0,0]$ 
  $C_{i,i,1}^{g,1} = D[i,g,1,1,0] + D^{\prime}[i,1,1,0]$ \newline
\bfseries{for} $w=1...(n-1)$ \newline
 \hspace*{1em} \bfseries{for} $i = 1 ... (n-w)$\newline
 \hspace*{2em} $j = i + w$ \newline
 \hspace*{2em} \bfseries{for} $g < i$ \bfseries{or}  $g > j$ \newline
 \hspace*{3em} $I_{i,j,0}^{g,0} = \boldsymbol{\rm logsumexp}_{i \le r <j} \left\{ C_{i,r,1}^{0,1} + C_{r+1,j,0}^{i,0} + D[j,g,0,0,1] + D^{\prime}[j,0,0,1] + S[j,i,g,0,0] + S^{\prime}[j,i,0,0] \right\} $ \newline
  \hspace*{3em} $I_{i,j,0}^{g,1} = \boldsymbol{\rm logsumexp}_{i \le r <j} \left\{ C_{i,r,1}^{0,1} + C_{r+1,j,0}^{i,0} + D[j,g,0,1,1] + D^{\prime}[j,0,1,1] + S[j,i,g,0,1] + S^{\prime}[j,i,0,1]  \right\} $ \newline
   \hspace*{3em} $I_{i,j,1}^{g,0} = \boldsymbol{\rm logsumexp}_{i \le r <j} \left\{ C_{i,r,1}^{j,0} + C_{r+1,j,0}^{0, 1} + D[i,g,1,0,1] + D^{\prime}[i,1,0,1] + S[i,j,g,1,0] + S^{\prime}[i,j,1,0] \right\} $ \newline
  \hspace*{3em} $I_{i,j,1}^{g,1} = \boldsymbol{\rm logsumexp}_{i \le r <j} \left\{ C_{i,r,1}^{j,0} + C_{r+1,j,0}^{0,1} + D[i,g,1,1,1] + D^{\prime}[i,1,1,1] + S[i,j,g,1,1] + S^{\prime}[i,j,1,1] \right\} $ \newline
     \hspace*{3em} $C_{i,j,0}^{g,0} = \boldsymbol{\rm logsumexp}_{i \le r \le j} \left\{ C_{i,r,0}^{0,1} + I_{r,j,0}^{i,0} \right\} $ \newline
     \hspace*{3em} $C_{i,j,0}^{g,1} = \boldsymbol{\rm logsumexp}_{i \le r \le j} \left\{ C_{i,r,0}^{0,1} + I_{r,j,0}^{i,1}  \right\} $ \newline
     \hspace*{3em} $C_{i,j,1}^{g,0} = \boldsymbol{\rm logsumexp}_{i \le r \le j} \left\{ I_{i,r,1}^{j,0} + C_{r,j,1}^{0,1} \right\} $ \newline
  \hspace*{3em} $C_{i,j,1}^{g,1} = \boldsymbol{\rm logsumexp}_{i \le r \le j} \left\{ I_{i,r,1}^{j,1} + C_{r,j,1}^{0,1}  \right\} $ \newline
\bfseries{for} $i = 1 .. (n)$ \newline
\hspace*{1em} $P[i] = R[i] + R^{\prime}[i] + C_{1,i,0}^{0,1}$ \newline
$P = \boldsymbol{\rm logsumexp}_{1\le i \le n} P[i] + C_{i,n,1}^{0,1}$ \newline
\Return P
 \caption{Inside algorithm for joint L-NDMV and sibling-NDMV} 
 \label{eisner joint}
\end{algorithm}

\end{document}